\documentclass[10pt, a4paper]{article}

\usepackage{lrec-coling2024} 
\usepackage{algorithmic}

\usepackage{amsmath}
\usepackage{amssymb}

\usepackage{multirow}
\usepackage{float}

\usepackage{subfig}
\usepackage{makecell}

\newcommand \footnoteONLYtext[1]
{
	\let \mybackup \thefootnote
	\let \thefootnote \relax
	\footnotetext{#1}
	\let \thefootnote \mybackup
	\let \mybackup \imareallyundefinedcommand
}

\title{Dual Encoder: Exploiting the Potential of Syntactic and Semantic for Aspect Sentiment Triplet Extraction\\ \vspace*{.5\baselineskip}}

\name{Xiaowei Zhao, Yong Zhou, Xiujuan Xu}

\address{School of Software Technology, Dalian University of Technology \\
         xiaowei.zhao@dlut.edu.cn, 89@mail.dlut.edu.cn, xjxu@dlut.edu.cn}

\abstract{
Aspect Sentiment Triple Extraction (ASTE) is an emerging task in fine-grained sentiment analysis. Recent studies have employed Graph Neural Networks (GNN) to model the syntax-semantic relationships inherent in triplet elements. However, they have yet to fully tap into the vast potential of syntactic and semantic information within the ASTE task. In this work, we propose a \emph{Dual Encoder: Exploiting the potential of Syntactic and Semantic} model (D2E2S), which maximizes the syntactic and semantic relationships among words. Specifically, our model utilizes a dual-channel encoder with a BERT channel to capture semantic information, and an enhanced LSTM channel for comprehensive syntactic information capture. Subsequently, we introduce the heterogeneous feature interaction module to capture intricate interactions between dependency syntax and attention semantics, and to dynamically select vital nodes. We leverage the synergy of these modules to harness the significant potential of syntactic and semantic information in ASTE tasks. Testing on public benchmarks, our D2E2S model surpasses the current state-of-the-art(SOTA), demonstrating its effectiveness.
 \\ \newline  }
 
\begin{document}

\maketitleabstract

\section{Introduction}

\begin{figure}[t]
    \centering
    \includegraphics[width=1.0\columnwidth]{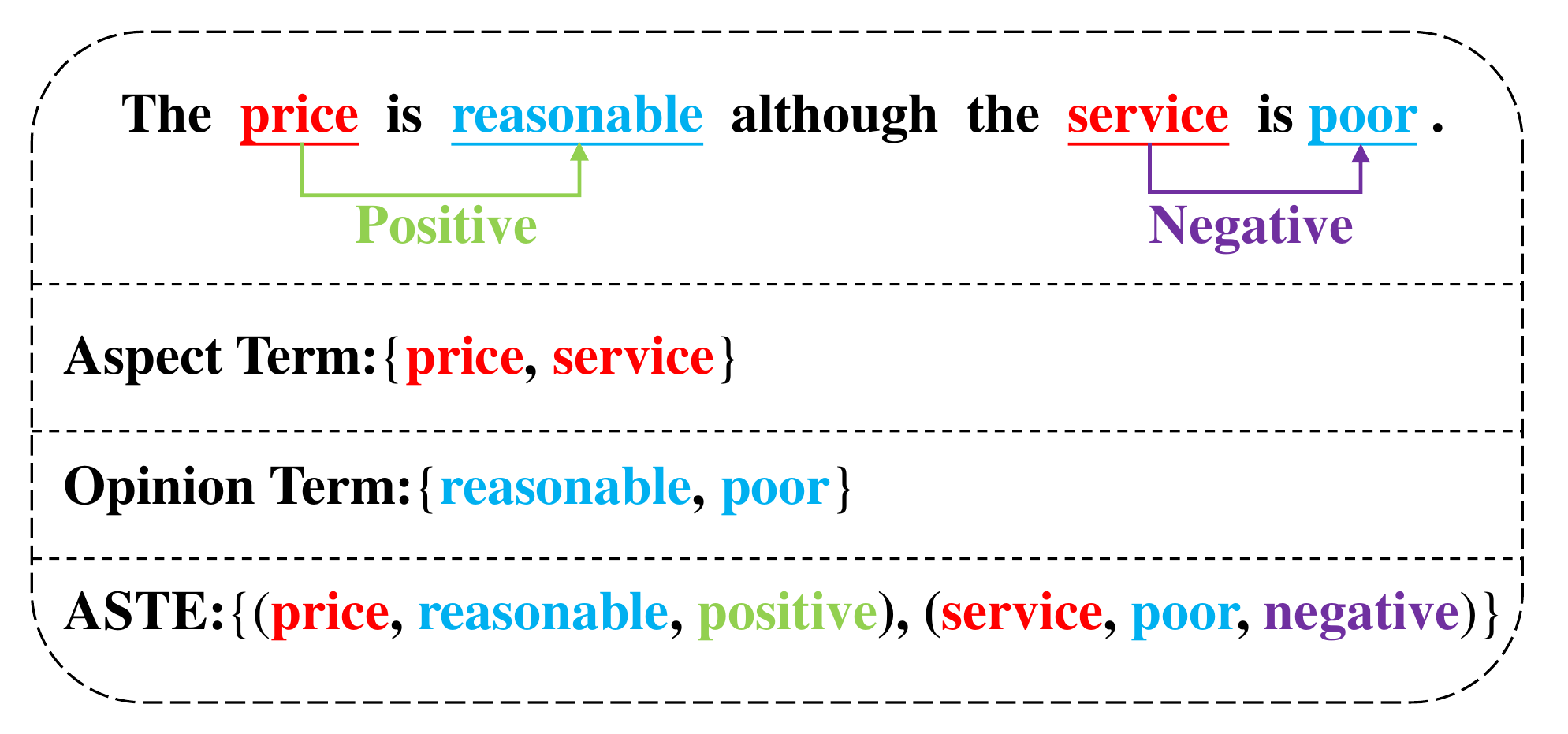}
    \caption{
    An example of the ASTE task. Aspect terms and opinion terms are highlighted in red and blue, respectively. Positive sentiment polarity is denoted by the color green, while purple symbolizes negative sentiment polarity.
    }
    \label{triplets}
\end{figure}

Aspect Sentiment Triplet Extraction (ASTE) is an advanced natural language processing task. In contrast to traditional sentiment analysis tasks, ASTE specifically targets fine-grained sentiment and aspectual information. It's objective is to identify aspect terms, opinion terms, and their associated sentiment in a sentence, as exemplified by the triplets (\emph{price, reasonable, positive}) and (\emph{service, poor, negative}) in Figure \ref{triplets}.

The ASTE task was initially introduced by \citet{DBLP:conf/aaai/PengXBHLS20}, they proposed a two-stage pipeline method for extracting triplets. However, this pipeline approach breaks the triplet structure's interactions and generally suffers from error propagation. \citet{DBLP:journals/corr/abs-2010-04640} proposed a novel grid labeling scheme (GTS), which transforms opinion pair extraction into a unified grid labeling task to solve the pipeline error propagation problem in an end-to-end manner. Such end-to-end solutions \cite{DBLP:conf/emnlp/WuWP20, DBLP:conf/emnlp/XuLLB20} heavily rely on word-to-word interactions to predict sentiment relations, ignoring semantics and syntactic relations between different spans. \citet{DBLP:conf/acl/ChenHLSJ21} propose a semantic and syntactic enhanced ASTE model ($\text{S}^{3}\text{E}^{2}$), which syntactic dependencies, semantic associations, and positional relationships between words are integrated and encoded into a graph neural network(GNN). Although their work has produced excellent results, we still believe that the model is far from realizing the strong potential brought by syntactic and semantic features for the ASTE task.

Previous studies \cite{DBLP:conf/acl/LiCFMWH20, DBLP:conf/acl/ChenHLSJ21, DBLP:conf/naacl/ZhangZW22} typically utilize one of BERT or LSTM to simultaneously extract syntactic and semantic features. However, a single encoder tends to specialize in either grammatical rules or semantic relationships, with a preference for one over the other. This segregated approach may result in partial and omitted information, particularly when dealing with complex or ambiguous sentences. So none of these models are able to realize the syntactic potential of syntactic and semantic features.

To fully exploit the enormous potential of syntactic and semantic information, we introduce the model \textbf{D}ual \textbf{E}ncoder: \textbf{E}xploiting the potential of \textbf{S}yntactic and \textbf{S}emantic (D2E2S), designed specifically for the ASTE task. 

\textbf{Firstly}, we selected BERT as the first encoder due to its superior ability to capture semantic information among words. An enhanced LSTM is employed as the second encoder, which includes a combination of BERT, BiLSTM, and Self-Attention. This combined encoder is able to better capture the local dependencies and sequence information between words while overcoming the limitations of LSTM in modeling long-distance dependencies. The enhanced LSTM can effectively capture rich syntactic information. To provide a clearer illustration, we consider the dual encoders as dual channels, namely BERT channels and LSTM channels.

\textbf{Secondly}, we introduce the Heterogeneous Feature Interaction Module (HFIM). In this module, we employ self-attention double-pooling (SADPool) to adaptively select crucial nodes from various perspectives. Simultaneously, through multiple rounds of information transfer via GCNConv and the selective neighbor information aggregation of GatedGraphConv, more distant neighbor information can be effectively conveyed and consolidated. The SADPool method is complemented by multi-layer GCNConv and GatedGraphConv to significantly enhance interactive performance, enabling the model to better filter and capture advanced syntactic and semantic information within the input features. GCNConv and GatedGraphConv correspond to the convolution calculations of the single-layer graph convolutional network \cite{DBLP:conf/iclr/KipfW17} and gated graph convolutional network \cite{DBLP:journals/corr/LiTBZ15}."

\textbf{Moreover}, the syntactic and semantic representations learned from SynGCN and SemGCN modules should show significant differences \cite{DBLP:conf/acl/LiCFMWH20}, we propose a strategy for separating syntactic and semantic similarity to enhance the model's ability to differentiate.

\textbf{In summary}, BERT encoders specialize in capturing semantic information between words, whereas enhanced LSTM encoders are more effective in capturing local dependencies, particularly dependency syntactic features. By employing a strategy that separates syntactic and semantic similarity, we have successfully obtained more distinctive syntactic and semantic information, while eliminating redundant interference, thereby enhancing the model's ability to differentiate between syntactic and semantic information. The SADPool method in the HFIM, in combination with multiple GCNConv and GatedGraphConv layers, more efficiently filters and captures advanced syntactic and semantic information within the input features. The syntax parser produces initial dependency syntactic features, while Multi-Head Attention (MHA) generates primary semantic features. These two types of raw syntactic and semantic features synergistically combine through the mentioned modules to fully exploit the significant potential of syntactic and semantic features.

Our contributions are highlighted as follows:

\textbf{1)} We propose dual encoders (BERT channel and enhanced LSTM channel) to enhance the representation of syntactic and semantic information.

\textbf{2)} We introduce the Heterogeneous Feature Interaction module (HFIM), where the SADPool technique synergizes with multi-layer GCNConv and GatedGraphConv, resulting in a significant enhancement of interactive performance. This collaboration enables the model to more effectively select and capture advanced syntactic and semantic information within the input features.

\textbf{3)} We present a strategy for separating syntactic and semantic similarities, enabling the model to effectively differentiate between syntactic and semantic information.

\textbf{4)} Our goal is to leverage the strengths of different modules to enhance the overall representation to fully harness the vast potential of syntactic and semantic features. We conduct comprehensive experiments on four benchmark datasets and surpass the current SOTA. Additionally, the source code and preprocessed datasets used in our work are provided on GitHub \footnote{\url{https://github.com/TYZY89/D2E2S}}.

\section{Related Work}

Aspect-Based Sentiment Analysis (ABSA) is a parominent research domain in the realm of Natural Language Processing (NLP). The ABSA tasks can be divided into single ABSA tasks and compound ABSA tasks\ \cite{DBLP:journals/corr/abs-2203-01054}.  The single ABSA task consists of multiple subtasks, such as aspect term extraction (ATE)\ \cite{DBLP:conf/emnlp/LiuJM15, DBLP:conf/acl/XuLSY18, DBLP:conf/coling/YangLQSS20, DBLP:conf/emnlp/WangWZYX21}, opinion term extraction (OTE)\ \cite{DBLP:conf/emnlp/LiL17, DBLP:conf/emnlp/WuWP20, DBLP:conf/emnlp/VeysehNDDN20, DBLP:conf/emnlp/MensahSA21} and so on. In contrast, single ABSA is more tightly coupled, whereas compound ABSA is more modular, allowing more flexibility in handling and improving each subtask, including aspect-opinion pair extraction (AOPE)\ \cite{DBLP:conf/acl/ZhaoHZLX20, DBLP:conf/acl/ChenLWZC20, DBLP:conf/aaai/GaoWLWZL21, DBLP:conf/ijcai/Wu0RJL21}, aspect category sentiment detection (ACSD)\ \cite{DBLP:conf/aaai/WanYDLQP20, DBLP:journals/kbs/WuXYYZGC21, DBLP:conf/acl/Zhang0DBL20} and so on. Nevertheless, none of these compound ABSA tasks focuses on extracting aspect terms along with their corresponding opinion terms and sentiment polarity in a unified manner. 

\citet{DBLP:conf/aaai/PengXBHLS20} initially introduced the Aspect Sentiment Triple Extraction(ASTE) task and proposed a two-stage pipeline method for extracting triplets. To further explore this task, \citet{DBLP:conf/emnlp/XuLLB20} proposed a location-aware labeling scheme that combines target locations and corresponding opinion spans to address the limitations of the BIOES labeling scheme. \citet{DBLP:journals/corr/abs-2010-04640} proposes a novel grid labeling scheme to solve the ASTE task in an end-to-end manner with only one unified grid labeling task. \citet{DBLP:conf/aaai/ChenWLW21} uses the framework of machine reading comprehension to ask and answer questions on the input text to achieve joint extraction to solve ASTE tasks. \citet{DBLP:journals/kbs/LiLLCZ22} proposed a span-shared joint extraction framework to simultaneously identify an aspect item and the corresponding opinion item and sentiment in the last step to avoid error propagation. \citet{DBLP:conf/acl/ChenZFLW22} utilizes a multi-channel graph to encode the relationship between words and introduces four types of linguistic features to enhance the GCN model. \citet{DBLP:conf/emnlp/MukherjeeKP023} proposed a novel multi-task approach for fine-tuning the obtained model weights by combining the base encoder-decoder model with two complementary modules: a tagging-based Opinion Term Detector and a regression-based Triplet Count Estimator.

\section{Methodology}

\begin{figure*} [t]
	\centering
	\includegraphics[width=2.0\columnwidth]{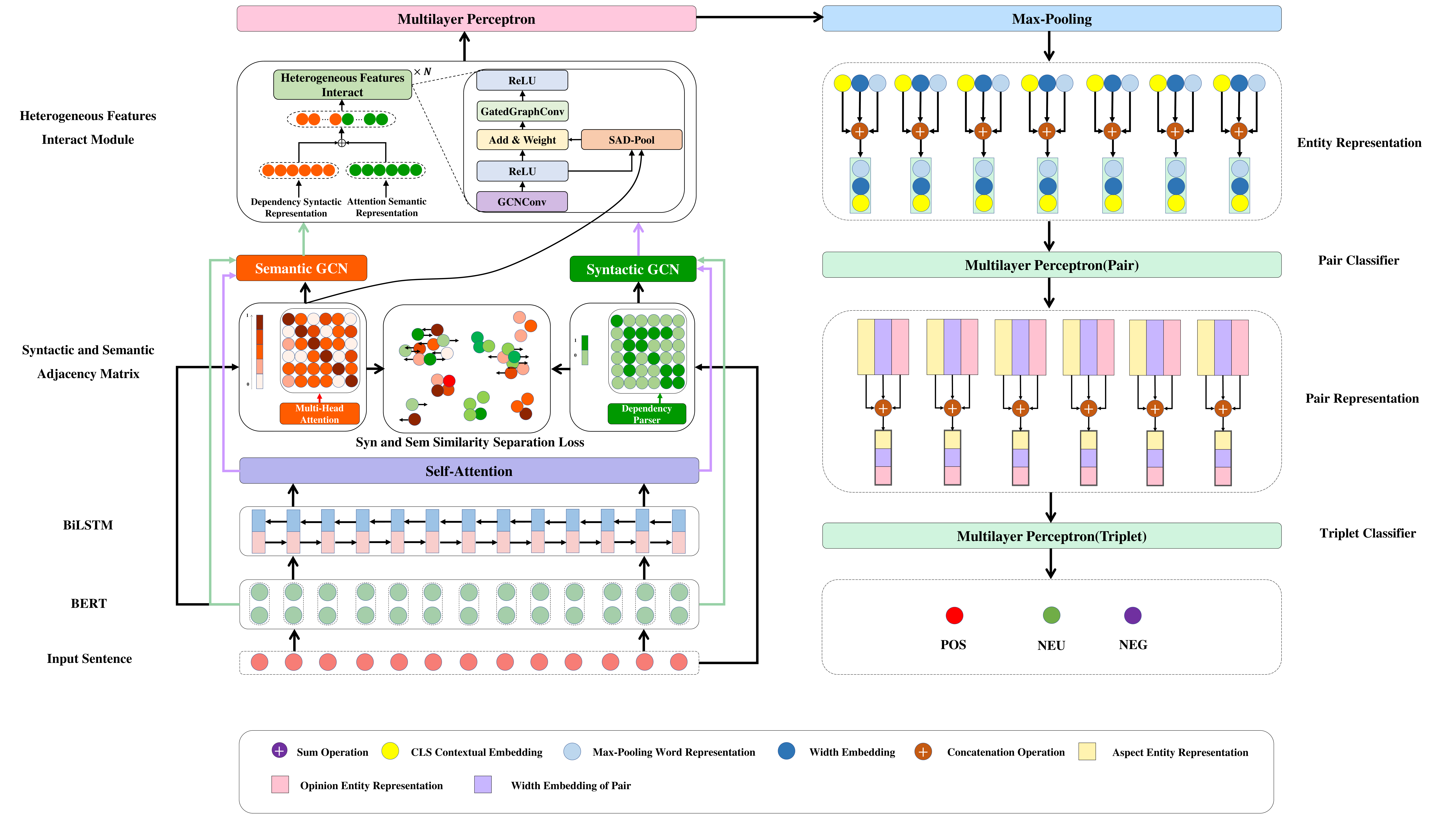}
	\caption{The overall architecture of our D2E2S model. The purple and light green arrows represent the LSTM and BERT channels respectively.}
	\label{framework}
\end{figure*}

The following sub-sections will explain the details of D2E2S. An overview of the D2E2S framework is shown in Figure \ref{framework}.

\subsection{Task Definition}

The Aspect Sentiment Triplet Extraction(ASTE) task aims to discern triplets $\mathcal{T}=\left\{{(a,o,s)}_{k}\right\}_{k=1}^{|\mathcal{T}|}$ within a sentence $X=\left\{w_{1},w_{2},\dots,w_{n}\right\}$ comprising $n$ tokens. Let $S$ be the set of all spans that can be enumerated, each triplet $(a,o,s)$ is defined as (aspect term, opinion term, sentiment polarity) where $s$ belongs to the set $\left\{Positive, Neutral, Negative\right\}$. 

\subsection{D2E2S Model}

\subsubsection{Input and Encoding Layer}

Dual encoders are employed to attain token-level contextual representations for a designated sentence X. The first encoder leverages traditional BERT for sentence feature extraction, while BERT-BiLSTM-SA(where SA denotes self-attention) is utilized as the second encoder for contextual representation extraction. For a more lucid illustration, we envisage the dual encoders as two distinct channels, specifically, the BERT channels and the LSTM channels. The encoding layer subsequently yields the hidden representation sequences $H^{lstm}=\left\{h_{1}^{lstm},h_{2}^{lstm},\dots,h_{n}^{lstm}\right\}$ and $H^{bert}=\left\{h_{1}^{bert},h_{2}^{bert},\dots,h_{n}^{bert}\right\}$ from the BERT-BiLSTM-SA Encoder and BERT Encoder, respectively. 

\subsubsection{Syntactic and Semantic Graph Convolutional Networks Construction}

\paragraph{SynGCN} 

To integrate syntactic information, we utilize the Stanford-NLP tool\footnote{\url{https://stanfordnlp.github.io/CoreNLP/}} for generating a syntactic dependency tree corresponding to the input sentence. We then build a bidirectional graph $G = (V, E)$ rooted in the dependency tree to encapsulate the syntactic relationships. The syntactic graph is embodied as an adjacency matrix $A^{syn} \in \mathbb{R}^{n \times n}$, which is defined as follows:
\begin{equation}
	\begin{split}
		A_{ij}^{syn}=\left\{\begin{array}{ll}
        1, & \text { if } x_{i}\text{ connect to} \ x_{j}\ \\
        0, & \text { otherwise }
        \end{array}\right.
	\end{split}
\end{equation}
As per this definition, $A_{ij}^{syn}$ signifies the element in the $i$-th row and $j$-th column of the adjacency matrix, which determines the presence of a syntactic link between nodes $x_i$ and $x_j$. Through the application of this adjacency matrix, the SynGCN module has the capacity to utilize syntactic information, thus augmenting the representation of spans.

\paragraph{SemGCN} 

For the creation of the attention score matrix $A^{sem}$, the Multi-Head Attention (MHA) mechanism is applied to the hidden state features $H_{ij}^{bert}$, derived from the BERT encoder. The MHA calculates the attention scores among words, with the softmax function being employed to normalize these scores. From a mathematical perspective, the attention score matrix $A^{sem}$ is formulated as follows:
\begin{equation}
    A^{sem}_{ij} = softmax(MHA(h_{i}^{bert}, h_{j}^{bert}))
\end{equation}
wherein the hidden state features $H^{bert}$, generated by the BERT channel, as the initial node representations within the semantic graph.

\subsubsection{Syntactic and Semantic Similarity Separation}

Similar syntactic and semantic distributions can intertwine, thereby influencing the overall context. As such, the model may require more comprehensive analysis and utilization of context information to overcome the challenges posed by this similarity. To mitigate this, \citet{DBLP:conf/acl/LiCFMWH20} utilize a differential regularizer between the two adjacency matrices, encouraging the SemGCN network to learn semantic features distinct from the syntactic features outlined by the SynGCN network. This approach leverages Euclidean distance for similarity-based separation, albeit with limited effectiveness. We propose a loss mechanism for syntactic and semantic adjacency matrices based on KL divergence, aiming to enhance the model's ability to distinguish between syntactic and semantic distributions more effectively. The loss of syntactic and semantic Similarity Separation is represented as follows:
\begin{equation}
  \begin{split}
    KL(A^{syn}_{i}||A^{sem}_{i}) =  \sum_{j}^{A^{sem}_{i}}f(A^{syn}_{i}) log \frac{f(A^{syn}_{i})}{f(A^{syn}_{j})}
  \end{split}
\end{equation}
\begin{equation}
  \begin{split}
    KL(A^{sem}_{i}||A^{syn}_{i}) =  \sum_{j}^{A^{sem}_{i}} f(A^{syn}_{j}) log \frac{f(A^{syn}_{j})}{f(A^{syn}_{i})}
  \end{split}
\end{equation}
\begin{equation}
  \begin{split}
    \mathcal{L}_{kl} &= \sum_{i}^{m} log\Bigl(1 +\\ &\bigl({|KL(A^{syn}_{i}||A^{sem}_{i})| + |KL(A^{sem}_{i}||A^{syn}_{i})|}\bigl)^{-1}\Bigl)
  \end{split}
\end{equation}
where f(·) represents softmax function.

\subsubsection{Heterogeneous Features Interact Module}

Prior work leveraged a Mutual BiAffine (Biaffine Attention) transformation for interaction between the SynGCN and SemGCN modules \cite{DBLP:conf/acl/LiCFMWH20}. In comparison, our heterogeneous features interact module, constituted by self-attention double-pooling(SADPool), multi-layer GCNConv and GatedGraphConv, demonstrates superior performance in modeling long-range dependencies and complex non-linear relationships within intricate contexts. Specifically, the purpose of SADPool is to accurately select essential nodes while mitigating the impact of non-essential ones. It achieves this by incorporating a self-attention mechanism for node scoring, GCNConv primarily updates node representations by leveraging neighbor node information from the graph structure. In contrast, GatedGraphConv concentrates on the use of node/edge features, introducing a gating mechanism. This mechanism adaptively filters and weights node/edge features with neighboring nodes' information, emphasizing vital features and relationships while disregarding irrelevant information. We utilize multi-layered GCNConv and GatedGraphConv to attain potent interactive performance. The implementation of multiple rounds of message passing and selective neighbor information aggregation facilitates superior filtering and capturing of high-level syntactic or semantic information present in the input features. 

\paragraph{Self-Attention Double-Pooling} 

Graph pooling methods today fall into two main categories: cluster pooling and top-k selection. Cluster pooling involves both structural and feature information, which can lead to assignment matrix issues. In top-k selection pooling, node importance is simplified, and unselected nodes lose their feature information, potentially resulting in significant graph information loss during pooling \cite{DBLP:conf/acl/KalchbrennerGB14, DBLP:conf/icml/LeeLK19, DBLP:conf/www/ZhangWL0SLLSB20}. Moreover, motivated by the work of MP-GCN \cite{DBLP:journals/access/ZhaoXW22}, we employ SADPool, utilizing the input attention adjacency matrix $A^{sem}$ of SemGCN as the input for SADPool. This adjacency matrix undergoes both average and max pooling from two distinct perspectives to select nodes with higher scores. To avoid substantial loss of graph information during pooling, we preserve all original graph data through the residual enhancement layer. The process of SADPool is formulated as follows:
\begin{equation}
  \begin{split}
    S_{mean } & =mean\left(A^{sem}\right), S_{max }=max \left(A^{sem}\right) 
  \end{split}
\end{equation}
\begin{equation}
  \begin{split}
    H^{(l+1)} & =\text{ReLU}\big(H^{(l)} \odot\left(1+S_{mean}+S_{max}\right)\big)
  \end{split}
\end{equation}
\begin{equation}
  \begin{split}
    H^{(l+1)} & =\text{SADPool}\big(H^{(l)}, A^{sem}\big)
  \end{split}
\end{equation}
Where $H^{(l)} \in \mathbb{R}^{N \times F}$ represents the output node representation at layer $l$, $N$ is the number of nodes and $F$ is the feature dimension for each node. $S_{mean} \in \mathbb{R}^{V \times 1}$ and $S_{max} \in \mathbb{R}^{V \times 1}$ are mean and maximum of $N$ groups of attention scores.

\paragraph{GCNConv}

The syntactic and semantic information output by the residual enhanced module is used as the initial node representation in GCNConv, and the specific update formula of GCNConv is as follows:
\begin{equation}
	\begin{split}
		H^{(l+1)}=\hat{D}^{-1 / 2} \hat{A} \hat{D}^{-1 / 2} H^{(l)} \Theta
	\end{split}
\end{equation}
\begin{equation}
	\begin{split}
		H^{(l+1)}=\text{GCNConv}\big(H^{(l)}, A\big)
	\end{split}
\end{equation}
Where $\hat{D}_{i i}\in \mathbb{R}^{N \times N}=\sum_{j=0} \hat{A}_{i j}$ represents the diagonal degree matrix, while $\hat{A}\in \mathbb{R}^{N \times N}=A+I$ signifies the adjacency matrix with self-loops included. Additionally, $\Theta \in \mathbb{R}^{F{in} \times F_{out}}$ denotes the parameters for the linear transformation of features, where $F_{in}$ and $F_{out}$ represent the dimensions of the input and output features, respectively.

\paragraph{GatedGraphConv}

The aggregated information of node $i$, denoted as $\mathbf{m}_{i}^{(l+1)}$, is obtained by performing a linear transformation between the adjacency matrix weight $e_{j, i}$ and the hidden state $\mathbf{h}_{j}^{(l)}$ of node $j$. This transformation is regulated by the parameter matrix $\boldsymbol{\Theta}$. In this context, the Gated Recurrent Unit (GRU) aggregates the node $i$'s information into $\mathbf{m}_{i}^{(l+1)}$ and takes the previous layer's hidden state $\mathbf{h}_{i}^{(l)}$ as input. The resulting output for node $i$ is the current layer's hidden state $\mathbf{h}_{i}^{(l+1)}$. The process of GatedGraphConv is formulated as follows:
\begin{equation}
	\begin{split}
		\mathbf{m}_{i}^{(l+1)} & =\sum_{j \in \mathcal{N}(i)} e_{j, i} \cdot \boldsymbol{\Theta} \cdot \mathbf{h}_{j}^{(l)}
	\end{split}
\end{equation}
\begin{equation}
	\begin{split}
		\mathbf{h}_{i}^{(l+1)} & =\text{GRU}\big(\mathbf{m}_{i}^{(l+1)}, \mathbf{h}_{i}^{(l)}\big)
	\end{split}
\end{equation}
\begin{equation}
	\begin{split}
		{H}^{(l+1)} =\text{GatedGConv}\big({H}^{(l)}\big)
	\end{split}
\end{equation}

\paragraph{Heterogeneous Features Interact}

The process for heterogeneous features interactinteractionis formulated as follows:
\begin{equation}
	\begin{split}
		[H^{lstm(syn)}, H^{bert(syn)}] &=\\ \text{SynGCN}(&H^{lstm}, H^{bert})
	\end{split}
\end{equation}
\begin{equation}
	\begin{split}
		H^{syn} = H^{lstm(syn)}\oplus H^{bert(syn)}
	\end{split}
\end{equation}
\begin{equation}
\begin{split}
        \widetilde{H}^{syn} = \sigma\Bigl(\text{GCNConv}(
        H^{syn}, A^{syn}, E^{index})\Bigl)
\end{split}
\end{equation}
\begin{equation}
\begin{split}
        \widetilde{H}^{syn^{p}} = \text{SADPool}\left(\widetilde{H}^{syn}, A^{sem}\right)
\end{split}
\end{equation}
\begin{equation}
	\begin{split}
		\Bigl[\widetilde{H}^{lstm(syn)}, \widetilde{H}^{bert(syn)}\Bigl] &= 
        \sigma\Bigl(\\\text{GatedGConv}(\widetilde{H}^{syn^{p}},&E^{index},E^{attr})\Bigl)
	\end{split}
\end{equation}
The term $E^{index}$(edge index) is generated by transforming the dense matrix to a sparse one, and the term $E^{attr}$(edge weight) is calculated based on the cosine similarity between the nodes, $\sigma$ is a nonlinear activation function (e.g., ReLU). Correspondingly, the terms $\widetilde{H}^{lstm(sem)}$ and $\widetilde{H}^{bert(sem)}$ in the formula can be expressed as follows:
\begin{equation}
	\begin{split}
        \Bigl[\widetilde{H}^{lstm(sem)},\widetilde{H}^{bert(sem)}\Bigl] &= 
        \sigma\Bigl(\\\text{GatedGConv}(\widetilde{H}^{sem^{p}}, &E^{index}, E^{attr})\Bigl)
	\end{split}
\end{equation}
We extract syntactic information from the LSTM channel and semantic information from the BERT channel, then employ a Multilayer Perceptron (MLP) to integrate these features. 
\begin{equation}
	\begin{split}
		\widetilde{H}^{out} = 
        \text{MLP}(\widetilde{H}^{lstm(syn)} \oplus \widetilde{H}^{bert(sem)})
	\end{split}
\end{equation}

\subsubsection{Dual Encoder Enhanced Syntactic and Semantic}

The two-channel encoder's output sequentially passes through the SynGCN (SemGCN) module, heterogeneous features interact module. Our goal is to leverage the strengths of different modules to enhance the overall representation to fully harness the vast potential of syntactic and semantic features, particularly enhancing the LSTM channel's syntactic information and the BERT channel's semantic information. The Multilayer Perceptron(MLP) module then fuses the syntactic and semantic feature information, yielding the final word representation. The resulting word representation seamlessly integrates extensive syntactic and semantic information while retaining the original feature information.

\subsubsection{Entity Representation and Filter}

We perform a series of operations to obtain a token-level contextualized representation $h_{i}^{out}$ from $\widetilde{H}^{out}$, which exhibits enhanced syntactic and semantic features and retains a higher proportion of the original encoded information of a given sentence.
\begin{equation}
	\begin{split}
		\mathbf{g}_{ij}=Max\left(h_{start}^{out}, h_{start+1}^{out}, \ldots, h_{e n d}^{out}\right)
	\end{split}
\end{equation}
\begin{equation}
	\begin{split}
		s_{ij}=g_{ij}\oplus f_{width}^{s}(i,j) \oplus h_{[cls]}
	\end{split}
\end{equation}
where $Max$ represents max pooling operation, and $f_{width}^{s}(i,j)$ denotes a trainable feature embedding that depends on the distance between the starting and ending indices of the span. The $h_{[cls]}$ feature, a component of the original coding feature, encapsulates specific global semantic information, acting as an augmentation to the contextual information across the span \cite{DBLP:journals/kbs/LiLLCZ22}. The representation of each enumerated span $s_{i,j} \in S$ serves as input for predicting the mention types $m \in\{ Target, Opinion \}$.
\begin{equation}
	\begin{split}
		P\left(m \mid s_{i, j}\right)=softmax\bigl(\text{MLP}_{m}\left(s_{i, j}\right)\bigl)
	\end{split}
\end{equation}
Consequently, the loss function of the span filter can be expressed by calculating the cross-entropy loss between the predicted distribution $P\left(m \mid s_{i, j}\right)$ and gold distribution $P\left(m^{\mathcal{T}} \mid s_{i, j}\right)$.
\begin{equation}
	\begin{split}
		\mathcal{L}_{s p}=-\sum_{s_{ij}\in S}P\left(m^{\mathcal{T}} \mid s_{i, j}\right) \log \left(P\left(m \mid s_{i, j}\right)\right)
	\end{split}
\end{equation}

\subsubsection{Sentiment Classifier}

In order to obtain the specific pair representation, we concatenate the aspect span $s_{a, b}^{t}$, the opinion span $s_{c, d}^{o}$, the trainable width feature embedding $f_{width}^{p}$, and the pair's context feature complement, $h_{[cls]}$.
\begin{equation}
	\begin{split}
		T_{s_{a, b}^{t},s_{c, d}^{o}}=s_{a, b}^{t} \oplus f_{width}^{p}\oplus  h_{[cls]}\oplus  s_{c, d}^{o}
	\end{split}
\end{equation}
\begin{equation}
	\begin{split}
		P\left(r \mid s_{a, b}^{t}, s_{c, d}^{o}\right)=softmax \Big(\text{MLP}_{r}\big(T_{s_{a, b}^{t},s_{c, d}^{o}}\big)\Big)
	\end{split}
\end{equation}
We feed the span pair representation $T$ into MLP layer, which determines the probability of sentiment relation $r \in R = \left\{Positive, Negative, Neutral \right\}$ between the target and the opinion. The sentiment classifier's loss function is computed by comparing the cross-entropy loss between the predicted distribution $P\left(r \mid s_{a, b}^{t}, s_{c, d}^{o}\right)$ and the gold distribution $P\left(r^{\mathcal{T}} \mid s_{a, b}^{t}, s_{c, d}^{o}\right)$.
\begin{equation}
	\begin{split}
		\mathcal{L}_{tri}&=- \sum_{s_{a, b}^{t} \in S^{t}, s_{c, d}^{o} \in S^{o}} \\&P\left(r^{\mathcal{T}} \mid s_{a, b}^{t}, s_{c, d}^{o}\right)\log P\left(r \mid s_{a, b}^{t}, s_{c, d}^{o}\right)
	\end{split}
\end{equation}

\subsubsection{Loss Function}

\begin{table}[t]
	\scriptsize
	\centering
	\begin{tabular}{cc|cccccc}
		\hline
		\multicolumn{2}{c|}{\multirow{2}{*}{Datasets}}      & \multirow{2}{*}{NEU} & \multirow{2}{*}{POS} & \multirow{2}{*}{NEG} & \multirow{2}{*}{\#S} & \multirow{2}{*}{\#T} \\
		\multicolumn{2}{c|}{}                               &                           &                           &                          &                           &                              &                            \\ \hline
		\multicolumn{1}{c|}{\multirow{3}{*}{14LAP}} & Train & 126                       & 817                       & 517                      & 906                       & 1460\\
		\multicolumn{1}{c|}{}                       & Dev   & 36                        & 169                       & 141                      & 219                       & 346\\
		\multicolumn{1}{c|}{}                       & Test  & 63                        & 364                       & 116                      & 328                       & 543\\ \hline
		\multicolumn{1}{c|}{\multirow{3}{*}{14RES}} & Train & 166                       & 1692                      & 480                      & 1266                      & 2338\\
		\multicolumn{1}{c|}{}                       & Dev   & 54                        & 404                       & 119                      & 310                       & 577\\
		\multicolumn{1}{c|}{}                       & Test  & 66                        & 773                       & 155                      & 492                       & 994\\ \hline
		\multicolumn{1}{c|}{\multirow{3}{*}{15RES}} & Train & 25                        & 783                       & 205                      & 605                       & 1013\\
		\multicolumn{1}{c|}{}                       & Dev   & 11                        & 185                       & 53                       & 148                       & 249\\
		\multicolumn{1}{c|}{}                       & Test  & 25                        & 317                       & 143                      & 322                       & 485\\ \hline
		\multicolumn{1}{c|}{\multirow{3}{*}{16RES}} & Train & 50                        & 1015                      & 329                      & 857                       & 1394\\
		\multicolumn{1}{c|}{}                       & Dev   & 11                        & 252                       & 76                       & 210                       & 339\\
		\multicolumn{1}{c|}{}                       & Test  & 29                        & 407                       & 78                       & 326                       & 514\\ \hline
	\end{tabular}
	\caption{Statistics for the ASTE-Data-V2 dataset. The abbreviations `NEU', `POS', and `NEG' stand for the counts of neutral, positive, and negative triplets, respectively. Similarly, `\#S' and `\#T' denote the number of sentences and triplets, respectively.}
	\label{datasets}
\end{table}
The model's overall loss objective is to minimize the following loss function: 
\begin{equation} 
    \begin{split} 
    \mathcal{L}=\mathcal{L}_{sp}+\mathcal{L}_{tri}+\alpha\mathcal{L}_{kl} \end{split} 
\end{equation}
The coefficients $\alpha$ are employed to control the impact of the associated relation constraint loss. These individual loss functions are then combined to form the comprehensive loss objective of the model.

\section{Experiments}

\begin{table*}[h]
	\footnotesize
	\centering
        \renewcommand{\arraystretch}{1.2} 
	\resizebox{\textwidth}{!}{\begin{tabular}{c|ccc|ccc|ccc|ccc}
			\hline
			& \multicolumn{3}{c|}{\textbf{14LAP}}                       & \multicolumn{3}{c|}{\textbf{14RES}}               & \multicolumn{3}{c|}{\textbf{15RES}}                       & \multicolumn{3}{c}{\textbf{16RES}}                        \\ \cline{2-13} 
			& P              & R              & F1             & P      & R              & F1             & P              & R              & F1             & P              & R              & F1             \\ \hline
               $\text{BARTABSA}^{*}$          & 61.41          & 56.19          &  58.69         &  65.52   & 64.99          &  65.25          & 59.14          & 59.38          & 59.26          & 66.6          & 68.68         &  67.62         \\
                $\text{GTS-BERT}^{\natural}$       & 57.52          & 51.92          & 54.58
               &70.92   &69.49           & 70.20          & 59.29          &
            58.07          & 58.67          & 68.58          & 66.60       
              & 67.58          \\
                $\text{Dual-MRC}^{\natural}$       & 57.39          & 53.88          & 55.58
               &71.55   &69.14           & 70.32          & 63.78          &
            51.87          & 57.21          & 68.60          & 66.24       
              & 67.40          \\
			$\text{EMC-GCN}^{*}$        & 61.70          & 56.26          & 58.81          & 71.21  & 72.39          & 71.78          & 61.54          & 62.47          & 61.93          & 65.62          & 71.30        
              & 68.33          \\
                $\text{Span-ASTE}^{*}$      & 63.44          & 55.84          & 59.38          & 72.89  & 70.89          &
            71.85          & 62.18          &  64.45         & 63.27          & 69.45          & 71.17          & 70.26          \\
                $\text{GAS}^{*}$          & -          & -          &  60.78         &  -   & -          &  72.16          & -          & -          & 62.10          & -          & -         &  70.10         \\
                $\text{SSJE}^{*}$      & 67.43          & 54.71          & 60.41          & 73.12  & 71.43          &
            72.26          & 63.94          &  66.17         & 65.05          & 70.82          & 72.00          & 71.38          \\
                 $\text{SBN}^{*}$      & 65.68          & 59.88          & 62.65          & 76.36  & 72.43          &
            74.34          & 69.93          &  60.41         & 64.82          & 71.59          & 72.57          & 72.08          \\
			$\text{SyMux}^{*}$          & -          & -          &  60.11         &  -   & -          &  74.84          & -          & -          & 63.13          & -          & -         &  72.76         \\
                $\text{CONTRASTE}^{*}$          & 64.20          & 61.70          &  62.90         &  73.60  & 74.40          &  74.00          & 65.30          & 66.70          & \textbf{66.10}          & 72.22          & 76.30         &  74.20         \\
                $\text{RLI}^{*}$          & -          & -          &  61.97         &  -   & -          &  74.98          & -          & -          & 65.71          & -          & -         &   73.33         \\
			\textbf{D2E2S(Ours)} & 67.38 & 60.31 & \textbf{63.65} & 75.92  & 74.36 & \textbf{75.13} & 70.09 & 62.11          & 65.86  & 77.97          &71.77 & \textbf{74.74}
            \\ \hline
	\end{tabular}}
	\caption{These are the experimental results on $\mathcal{D}_{2}$ test datasets. The symbol $"{\natural}"$ indicates that the results have been retrieved from \citet{DBLP:conf/acl/ChenZFLW22}. The asterisk $"*"$ denotes that the results have been sourced from the original papers, with the highest scores highlighted in bold.}
	\label{main results}
\end{table*}

\subsection{Datasets}

We conducted an evaluation of our proposed model on four benchmark datasets $\mathcal{D}_{2}$\footnote{\url{https://bit.ly/3Ql5Yw0}} courtesy of \citetlanguageresource{DBLP:conf/emnlp/XuLLB20_lrec}. These sets are comprised of three restaurant-focused datasets and one laptop-focused dataset. The original ASTE datasets were published by \citetlanguageresource{DBLP:conf/aaai/PengXBHLS20_lrec}, \citetlanguageresource{DBLP:conf/emnlp/XuLLB20_lrec} refined them by remedying missing triplets and eliminating triplets with conflicting sentiments. For more details on these datasets, please refer to Table \ref{datasets}.

\subsection{Baselines}

We compare the proposed model with several leading benchmark methods. The following is a brief description of some of our selected benchmark methods:
\begin{itemize}
    \item\textbf{GTS}\ \cite{DBLP:journals/corr/abs-2010-04640} treats the task as a unified grid tagging task, employing an innovative tagging scheme for concurrent extraction of opinion triplets.
    \item\textbf{Span-ASTE}\ \cite{DBLP:conf/acl/XuCB20} constructs span representations for all potential target and opinion terms, with each possible target-opinion pair having its sentiment relation independently predicted.
    \item\textbf{EMC-GCN}\ \cite{DBLP:conf/acl/ChenZFLW22} employs multi-channel graph convolution operations to model diverse relation types between word pairs and leverages GCN to learn node representations that are aware of these relations.
    \item \textbf{SyMux}\ \cite{DBLP:conf/ijcai/0001LLWLJ22} introduces a novel multi-cascade framework that decomposes the sentiment analysis task into seven subtasks, fostering efficient interaction among these subtasks through multiple decoding mechanisms.
    \item \textbf{CONTRASTE}\ \cite{DBLP:conf/emnlp/MukherjeeKP023} proposed a novel multi-task approach for fine-tuning the obtained model weights by combining the base encoder-decoder model with two complementary modules: a tagging-based Opinion Term Detector and a regression-based Triplet Count Estimator.
    
\end{itemize}
Additionally, our comparisons also include models not described above, namely \textbf{BARTASA} \cite{DBLP:conf/acl/YanDJQ020}, \textbf{Dual-MRC} \cite{DBLP:conf/aaai/MaoSYC21}, \textbf{GAS} \cite{DBLP:conf/acl/Zhang0DBL20}, \textbf{SSJE} \cite{DBLP:journals/kbs/LiLLCZ22}, \textbf{SBN} \cite{DBLP:conf/emnlp/0015CSZ22} and \textbf{RLI} \cite{DBLP:conf/acl/YuLJ0023}. These models act as vital benchmarks, helping to provide a more comprehensive evaluation of our proposed method's performance.

\subsection{Implementation Details}

In our experiments, we employ the uncased base version of BERT \footnote{\url{https://huggingface.co/bert-base-uncased}}. We set the hidden state dimensionality to 768 for BERT and 384 for BiLSTM, with the dropout rate configured to 0.5. For BERT's fine-tuning, we leverage the AdamW optimizer \cite{DBLP:conf/iclr/LoshchilovH19} with a maximum learning rate of 5e-5 and a weight decay of 1e-2. The maximum span length is capped at 8. Regarding the architecture, SynGCN consists of 2 layers, and the same holds true for SemGCN. As for Heterogeneous Feature Interaction, we set the number of layers to 1. This configuration remains consistent across GCNConv, GatedGraphConv, and SADPool, where all have a single layer. The hyper-parameter $\alpha$ is 10. The training process is performed on Google Colab T4, lasting through 120 epochs with a mini-batch size of 16.

\subsection{Main Results}

\begin{table}[t]
	\footnotesize
	\centering
	\resizebox{\linewidth}{!}{
		\begin{tabular}{ccccc}
		  \hline
                \textbf{Model}     & \textbf{14lap}           & \textbf{14res} & \textbf{15res}         & \textbf{16res}              \\ \hline
			\multicolumn{1}{l}{D2E2S} & \textbf{63.65} & \textbf{75.13} & \textbf{65.86} & \textbf{74.74} \\
                \multicolumn{1}{l}{\quad {$\textbf{\textit{W/O}}$ SS}} & 60.49 & 71.04 & 63.61 & 71.98 \\
                \multicolumn{1}{l}{\quad {$\textbf{\textit{W/O}}$ Syntactic}} & 61.37 & 74.51 & 63.83 & 73.84 \\
                \multicolumn{1}{l}{\quad {$\textbf{\textit{W/O}}$ Semantic}} & 61.48 & 72.94 & 65.54 &  71.56 \\
                \multicolumn{1}{l}{\quad {$\textbf{\textit{W/O}}$ HFIM}} & 63.34 & 73.13 & 64.55 & 72.71 \\
                \multicolumn{1}{l}{\quad (E1+E2) $\rightarrow$ (E1)} & 61.91 & 73.39 & 61.91 & 73.11 \\ 
                \multicolumn{1}{l}{\quad (E1+E2) $\rightarrow$ (E2)} & 62.26 & 72.92 & 62.90 & 72.97 \\
			\multicolumn{1}{l}{\quad HFIM $\rightarrow$ Mutual BiAffine} & 62.55 & 71.60 & 64.25 & 71.69 \\ 
 \hline
		\end{tabular}
	}
	\caption{F1 scores of ablation study on $\mathcal{D}_{2}$ .}
	\label{Ablation results}
\end{table}

\begin{figure*}[t]
    \centering
    \includegraphics[width=2.0\columnwidth]{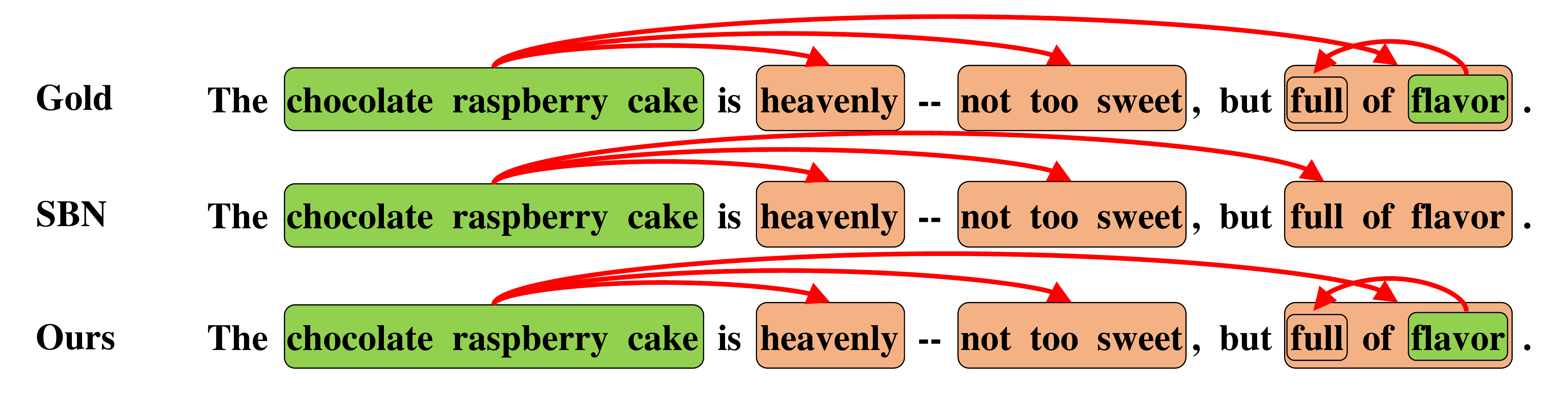}
    \caption{
    An example of the ASTE task. Aspect terms and opinion terms are highlighted in green and orange, respectively. Positive sentiment polarity is denoted by the color red.
    }
    \label{Case_Study}
\end{figure*}

The principal results are displayed in Table \ref{main results}. According to the F1 metric, our D2E2S model, bolstered by dual encoders, surpasses the current state-of-the-art (SOTA) \footnote{(\url{https://bit.ly/3Qj3Rc4}) as of October 15, 2023.} model, SyMux\ \cite{DBLP:conf/ijcai/0001LLWLJ22}, in the 14lap, 14res, 15res and 16res benchmarks by 3.54\%, 0.29\%, 2.73\%, and 1.98\% F1 points respectively. While the SOTA model amalgamates three versions of the ABSA dataset to glean additional labels for all 7 tasks, our model surpasses its performance using only a single version of the ABSA dataset.

\subsection{Model Analysis}

\subsubsection{Ablation Study}

We conducted ablation experiments to evaluate the effectiveness of various components of the D2E2S model. Table \ref{Ablation results} presents the experimental results. For simplicity, we denoted "BERT" as "E1", "BERT-BiLSTM-SA(Self-Attention)" as "E2", "Heterogeneous Features Interact Module" as "HFIM", and "Syntactic and Semantic" as "SS". 
\textbf{\emph{W/O}}  syntactic, semantic, and both syntactic and semantic features, performance decreased across all four datasets. This consistent trend indicates that both syntactic and semantic features play crucial roles in enhancing model performance. They complement each other, providing a solid foundation for successful ASTE task execution across these diverse datasets. \textbf{\emph{W/O}} HFIM, D2E2S's capacity to select and capture high-level syntactic and semantic information was hindered. Consequently, a slight performance reduction was observed on the 14lap datasets, with more noticeable drops of 2.00\%, 1.31\%, and 2.03\% on the 14res, 15res, and 16res datasets, respectively.  \textbf{\emph{Replacing}} the feature interaction method from HFIM to Mutual BiAffine, the long-range and nonlinear modeling capabilities of D2E2S were negatively impacted, leading to a significant performance degradation across all sub-datasets. \textbf{\emph{Replacing}} the dual encoder with a single encoder in either the BERT or LSTM channel, the effective exploitation of the syntactic and semantic potential of the ASTE task was obstructed, resulting in a performance decrease across all sub-datasets. In conclusion, every component of the D2E2S model plays a significant role in the overall performance of the ASTE task. 

\subsubsection{Case Study}

By analyzing a particularly challenging example, we probe into the competencies of our method. Figure \ref{Case_Study} presents the predictions made by SBN \cite{chen2022span} and D2E2S, highlighting the aspect terms and opinion terms marked in green and orange, respectively. In previous work, SBN demonstrates robust modeling abilities in addressing "one-to-many" and "many-to-one" challenges. However, it falls short in detecting the multiple relations of a word-pair, both "flavor" and "full" not only belong to the same opinion term (\emph{i.e.}, full of flavor), but they form a valid aspect-opinion pair as well, resulting in multiple relations of the word pair “flavor-full”, which also challenges the existing scheme. Conversely, our D2E2S excels in accurate identification, thanks to its ultra-long distance modeling capability and rich syntactic and semantic information incorporation.

\section{Conclusions}

In this study, we introduce the D2E2S architecture with dual encoders, designed to harness the vast potential of syntactic and semantic information for ASTE. We build dual encoders to separately model the syntactic structure and semantic information in each sentence. Next, we present a method to separate syntactic and semantic similarity, with the goal of assisting the model in better distinguishing between syntactic and semantic information. Furthermore, we introduce the Heterogeneous Feature Interaction Module, where the SADPool method is combined with multi-layer GCNConv and GatedGraphConv to greatly improve interactive capabilities, allowing the model to effectively filter and capture advanced syntactic and semantic information from the input features. By integrating across the mentioned modules, we fully harness the substantial potential of syntactic and semantic features. Experimental results demonstrate that our network outperforms baseline models significantly, achieving state-of-the-art (SOTA) results.

\section{Limitations}

Despite the promising results of our study, there are still some limitations that should be acknowledged. These limitations underscore areas needing future improvement and exploration. Firstly, our model's span length is set at 8, significantly reducing computing resources for span-level models. However, this restricts the capture of aspects or opinion terms with a span length exceeding 8. Future research could concentrate on more flexible options. Secondly, this article's Heterogeneous Features Interact Module leverages the stacking of GCNConv and GatedGraphConv to facilitate interaction. The incoming parameters of GCNConv and GatedGraphConv are edge indices, specifying each node's interaction range. Enabling more nodes to interact necessitates substantial computing resources. However, owing to limited computing resources in this experiment, the dense graph was transformed into a sparse graph to minimize the number of interacting nodes, inevitably reducing the interaction effect and experimental performance. Lastly, computational power and time constraints prevented us from exploring larger model architectures or conducting extensive hyperparameter tuning. We hope future studies will address these limitations, thereby enhancing the reliability and applicability of our proposed method.



\section{Bibliographical References}

\bibliographystyle{lrec-coling2024-natbib}
\bibliography{lrec-coling2024-example}

\section{Language Resource References}
\label{lr:ref}
\bibliographystylelanguageresource{lrec-coling2024-natbib}
\bibliographylanguageresource{languageresource}

\end{document}